\documentclass{article}
\usepackage{amsmath}
\usepackage[final]{neurips_2024}
\usepackage[utf8]{inputenc} 
\usepackage[T1]{fontenc}    
\usepackage{hyperref}       
\usepackage{url}            
\usepackage{booktabs}       
\usepackage{amsfonts}       
\usepackage{nicefrac}       
\usepackage{microtype}      
\usepackage{xcolor}         
\usepackage{graphicx}
\usepackage{array}
\usepackage{tabularx} 
\usepackage{subcaption}

\title{Interpretable Learning Dynamics in Unsupervised
Reinforcement Learning}

\author{%
  Shashwat Pandey \\
  Group Name  : Optimizer\\
  Department of Computer Science \\
  Arizona State University \\
  \texttt{spande90@asu.edu} \\
}

\begin{document}

\maketitle

\begin{abstract}
We present an interpretability framework for unsupervised reinforcement learning (URL) agents, aimed at understanding how intrinsic motivation shapes attention, behavior, and representation learning. We analyze five agents—DQN, RND, ICM, PPO, and a Transformer-RND variant—trained on procedurally generated environments, using Grad-CAM, Layer-wise Relevance Propagation (LRP), exploration metrics, and latent space clustering.

To capture how agents perceive and adapt over time, we introduce two metrics: \textit{attention diversity}, which measures the spatial breadth of focus, and \textit{attention change rate}, which quantifies temporal shifts in attention. Our findings show that curiosity-driven agents display broader, more dynamic attention and exploratory behavior than their extrinsically motivated counterparts. Among them, Transformer-RND combines wide attention, high exploration coverage, and compact, structured latent representations.

Our results highlight the influence of architectural inductive biases and training signals on internal agent dynamics. Beyond reward-centric evaluation, the proposed framework offers diagnostic tools to probe perception and abstraction in RL agents, enabling more interpretable and generalizable behavior.
\end{abstract}

\section{Introduction}

Deep reinforcement learning (RL) has achieved remarkable success in a variety of complex environments, including video games, robotic control, and large-scale simulations. However, the internal decision-making processes of these agents remain largely opaque. As RL systems become increasingly integrated into real-world applications—such as autonomous driving platforms like Waymo—the need for interpretable and trustworthy agents is more critical than ever. This project addresses that need by investigating how different types of RL agents allocate attention and build internal representations during training.

In this work, we conduct a comprehensive interpretability study across five RL agents: Deep Q-Network (DQN), Intrinsic Curiosity Module (ICM), Random Network Distillation (RND), Proximal Policy Optimization (PPO), and a Transformer-based PPO agent with RND. These agents span value-based and policy-gradient methods, and include both extrinsically and intrinsically motivated exploration strategies. Using gradient-based and relevance-based visualization techniques such as Grad-CAM and Layer-wise Relevance Propagation (LRP), alongside VAE-based latent representation analysis, we explore how each agent perceives and interacts with procedurally generated environments.

This study offers both qualitative and quantitative insights into attention evolution, exploration behavior, and representational structure across agents. Curiosity-driven agents (ICM, RND) are shown to develop more adaptive and structured visual attention, while value-based agents like DQN exhibit narrower and more static saliency. Additionally, we include VAE reconstruction visualizations to highlight how internal latent spaces differ in fidelity and focus.

\textbf{Contributions:}
\begin{itemize}
    \item A unified interpretability framework combining Grad-CAM, LRP, and VAE-based analysis to study RL agents.
    \item Comparative evaluation across five agents (DQN, ICM, RND, PPO, Transformer-RND) on the CoinRun environment from the Procgen benchmark.
    \item Novel inclusion of VAE reconstruction visualizations for evaluating latent space quality.
    \item A fully independent execution of all modeling, training, visualization, and analysis pipelines.
\end{itemize}

By bridging attention visualization and representation analysis, this work aims to move beyond performance metrics toward a deeper understanding of how and why RL agents behave as they do.

\section{Methodology}

\subsection{Agents and Environment}

We conduct our study using five reinforcement learning (RL) agents with varying exploration and learning strategies: Deep Q-Network (DQN), Intrinsic Curiosity Module (ICM), Random Network Distillation (RND), Proximal Policy Optimization (PPO), and a Transformer-based PPO agent with RND. These span both value-based (DQN) and policy-gradient (PPO variants) paradigms and include both extrinsically and intrinsically motivated approaches.

All agents are trained in the \textbf{CoinRun} environment from the Procgen benchmark, designed for generalization studies. Each agent interacts with 10 procedurally generated levels, with RGB frame inputs resized to $64 \times 64$ pixels. We use 8 parallel environments for faster sampling, a total of 1 million steps per agent, and a consistent device setup (CUDA-enabled GPUs).

\subsection{Training Configuration}

Common training parameters across agents include a discount factor $\gamma = 0.99$, buffer size of 200k transitions, and batch sizes ranging from 64 to 128. All agents are optimized using Adam. PPO and Transformer agents use rollout lengths of 256 and 512 respectively, with 4 update epochs and clipping parameter $\epsilon = 0.2$.

\subsection{Intrinsic Motivation Formulations}

\paragraph{ICM.} The Intrinsic Curiosity Module defines an intrinsic reward based on prediction error in a learned forward dynamics model:
\[
r_t^{\text{int}} = \frac{\beta}{2} \| \hat{\phi}(s_{t+1}) - \phi(s_{t+1}) \|_2^2,
\]
where $\phi(\cdot)$ is a learned encoder, and $\hat{\phi}$ is predicted using the agent's action and current state. The total ICM loss is:
\[
\mathcal{L}_{\text{ICM}} = \mathcal{L}_{\text{inv}} + \beta \mathcal{L}_{\text{fwd}},
\]
combining inverse and forward model objectives. We set $\beta = 0.015$ in our experiments.

\paragraph{RND.} Random Network Distillation uses a fixed randomly initialized target network $f_{\text{tgt}}$ and a trainable predictor $f_{\text{pred}}$. The intrinsic reward is:
\[
r_t^{\text{int}} = \| f_{\text{pred}}(s_t) - f_{\text{tgt}}(s_t) \|_2^2,
\]
with only the predictor trained using MSE loss. This encourages exploration of novel states.

\subsection{Transformer-based Policy Model}

Our Transformer-RND agent integrates a convolutional encoder with a 4-layer Transformer module (128-d embedding, 8 attention heads). Spatial tokens from the CNN are linearly projected and passed through the Transformer encoder, followed by linear actor and critic heads. We observe this architecture stabilizes long-horizon representation learning and supports meaningful attention evolution over training.

\subsection{Interpretability Framework}

To study how agents attend to their environment, we apply two complementary visual attribution methods:
\begin{itemize}
    \item \textbf{Grad-CAM}: generates coarse class-discriminative localization maps using gradients from convolutional layers.
    \item \textbf{Layer-wise Relevance Propagation (LRP)}: propagates relevance scores backward from outputs to input pixels for fine-grained interpretability.
\end{itemize}

We apply both techniques at five key training checkpoints (0k, 100k, 500k, 800k, 995k steps) for all agents. These are visualized and compared qualitatively in the next section.

\subsection{VAE-based Representation Analysis}

To investigate the structure of internal representations, we train a separate Variational Autoencoder (VAE) on RGB frames collected from each agent. The VAE consists of:
\begin{itemize}
    \item \textbf{Encoder:} 3 convolutional layers $\rightarrow$ flatten $\rightarrow$ FC layers to latent mean $\mu$ and log variance $\log \sigma^2$
    \item \textbf{Latent space:} 32-dimensional, sampled via reparameterization trick
    \item \textbf{Decoder:} FC $\rightarrow$ deconv layers reconstruct original image
\end{itemize}

The loss function is the standard ELBO:
\[
\mathcal{L}_{\text{VAE}} = \mathbb{E}_{q(z|x)}[\log p(x|z)] - D_{KL}[q(z|x)\|p(z)],
\]
where $x$ is the input frame and $z$ is the latent code.

We use this VAE for both qualitative (reconstruction fidelity) and quantitative (trajectory entropy, clustering) analysis of each agent's state distribution. Latent vectors are projected using t-SNE and UMAP, and clustering quality is evaluated via Silhouette, Davies-Bouldin, and Calinski-Harabasz scores.

\section{Experiments and Results}

\subsection{Training Performance}

\begin{figure}[h]
    \centering
    \begin{subfigure}{\linewidth}
        \includegraphics[width=\linewidth]{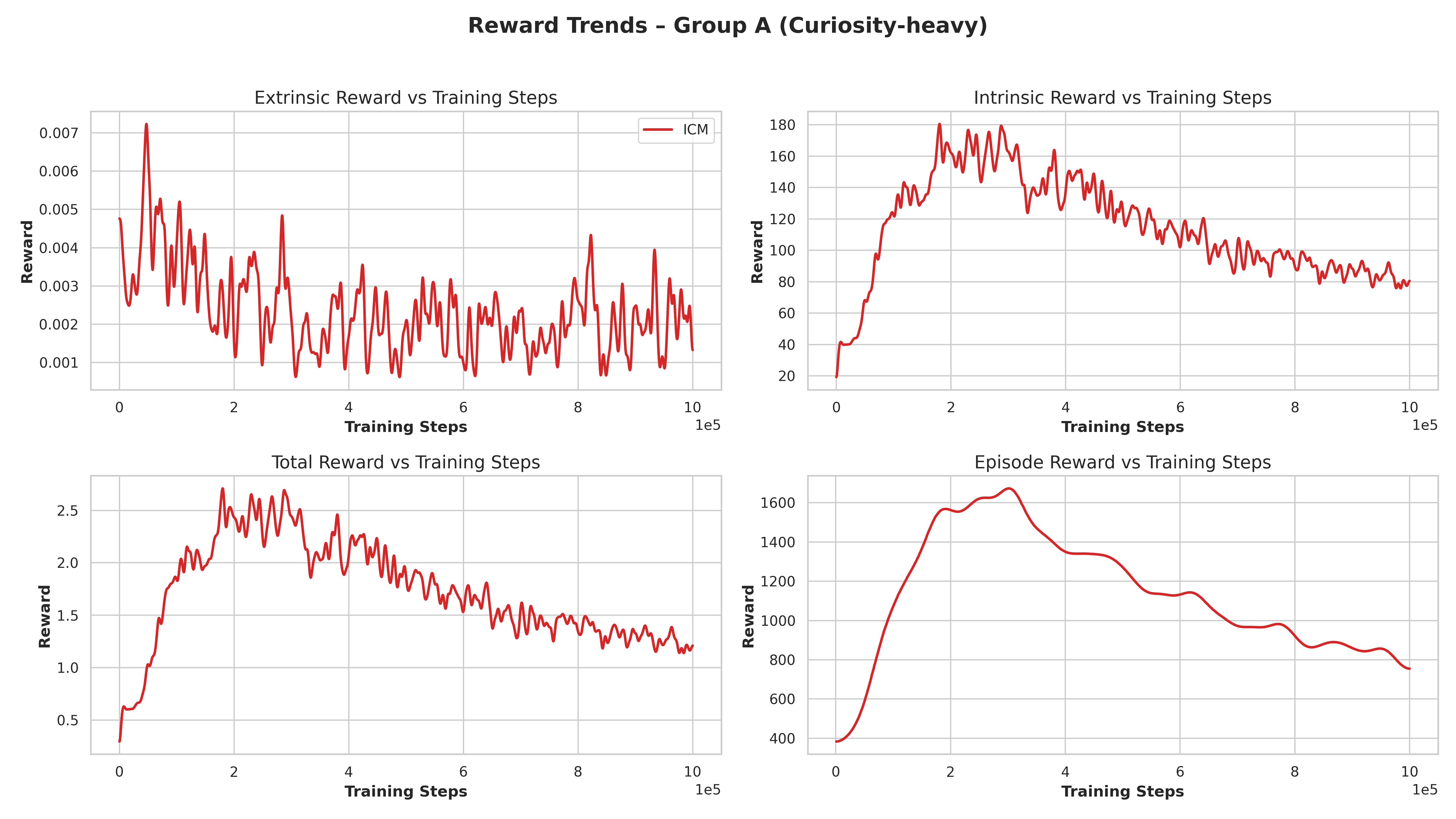}
        \caption{Group A: ICM}
    \end{subfigure}
    \hfill
    \begin{subfigure}{\linewidth}
        \includegraphics[width=\linewidth]{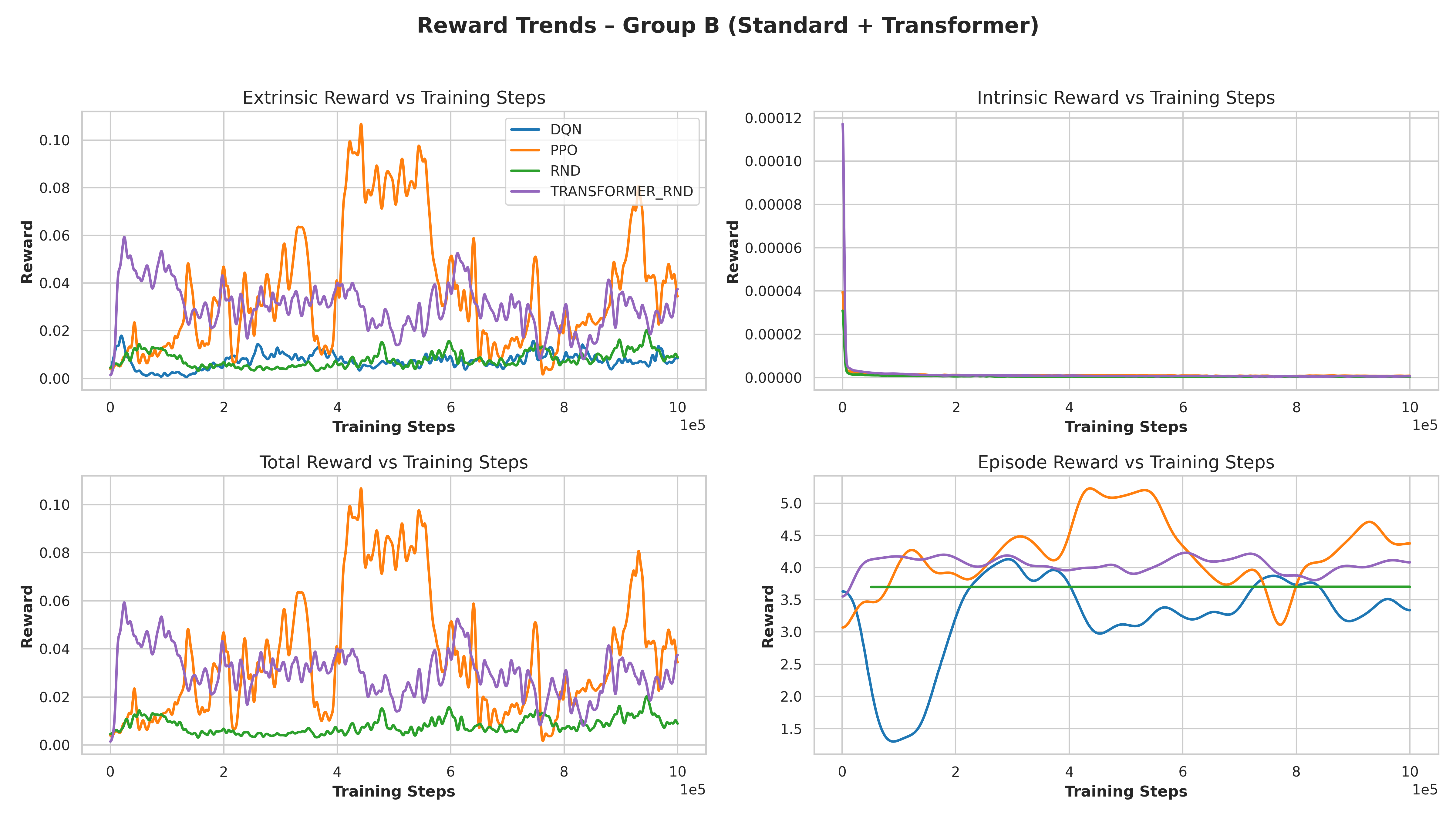}
        \caption{Group B: DQN, PPO, RND, Transformer-RND}
    \end{subfigure}
    \caption{Reward curves showing intrinsic, extrinsic, and episode returns for different agent groups.}
    \label{fig:reward_curves}
\end{figure}

We evaluate all agents on the CoinRun environment for 1 million training steps. Figure~\ref{fig:reward_curves} presents the reward dynamics for Group A (ICM) and Group B (DQN, PPO, RND, Transformer-RND), across intrinsic, extrinsic, total, and episode-level rewards.

Curiosity-driven agents like ICM show strong early intrinsic rewards, driving rapid exploration in the initial stages. However, this advantage tapers off over time as novel states diminish. Transformer-RND and PPO achieve the highest and most stable extrinsic and episode rewards, indicating robust policy learning. In contrast, DQN exhibits consistently low and flat reward curves, reflecting its difficulty in sparse-reward environments.

\subsection{Visual Attention Evolution}

\begin{figure}[h]
    \centering
    \includegraphics[width=0.9\linewidth]{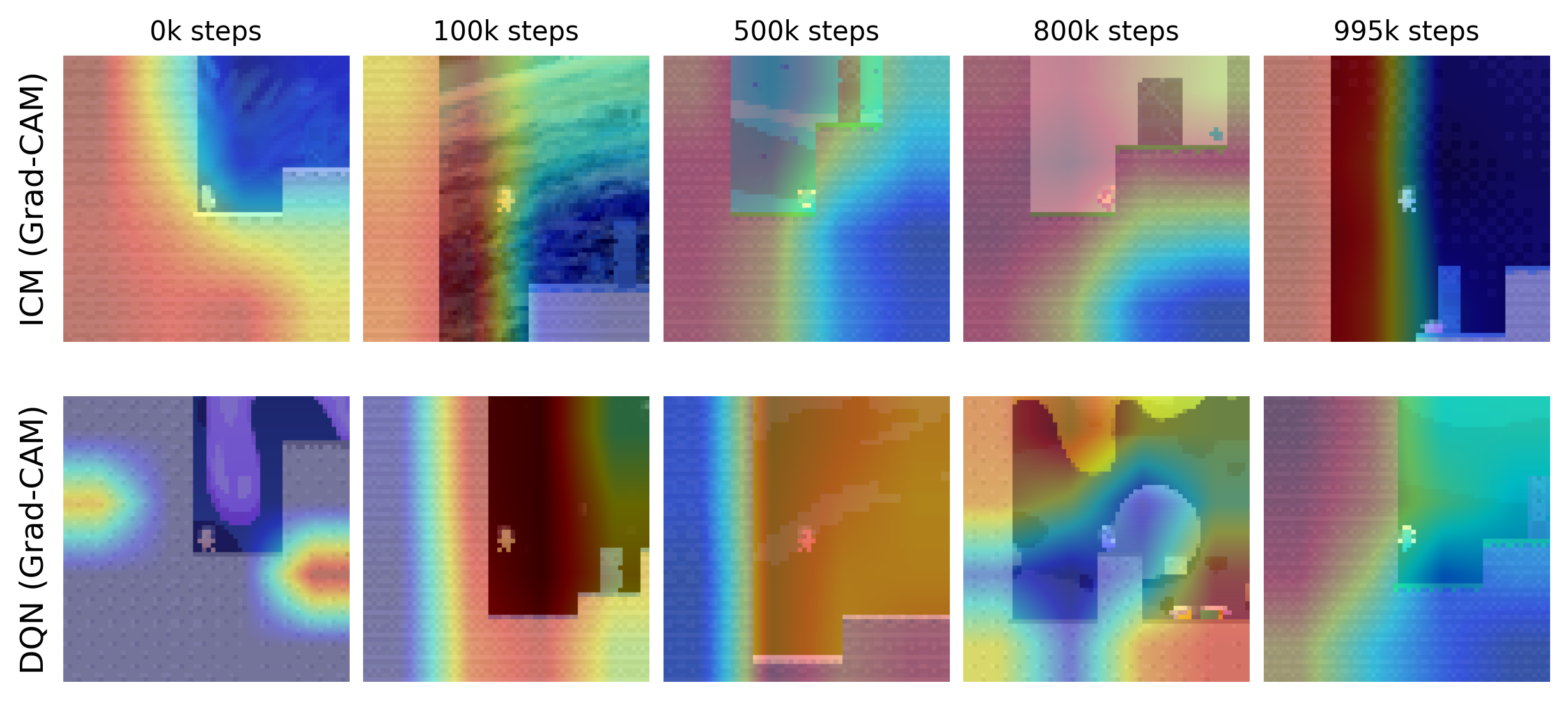}
    \caption{Grad-CAM visualizations over time for ICM (top) vs. DQN (bottom). ICM focuses on actionable areas, DQN remains noisy.}
    \label{fig:gradcam_comparison}
\end{figure}

We analyze Grad-CAM and Layer-wise Relevance Propagation (LRP) outputs at multiple training checkpoints. Figure~\ref{fig:gradcam_comparison} contrasts two agents: ICM (structured attention) vs. DQN (diffuse attention). We focus on ICM and DQN for visual comparison due to their clear contrast: ICM demonstrates the most structured and evolving attention maps, while DQN shows minimal progression and diffuse saliency. This stark difference makes them ideal for illustrating the spectrum of attention learning in reinforcement learning agents.

ICM gradually focuses on semantically meaningful regions (platform edges, collectibles), revealing increasingly precise saliency over training. In contrast, DQN exhibits inconsistent, broad, and often misplaced activations throughout, indicating weak spatial grounding of its policy.

\subsection{Representation Quality via VAE}

\begin{figure}[h]
    \centering
    \includegraphics[width=0.95\linewidth]{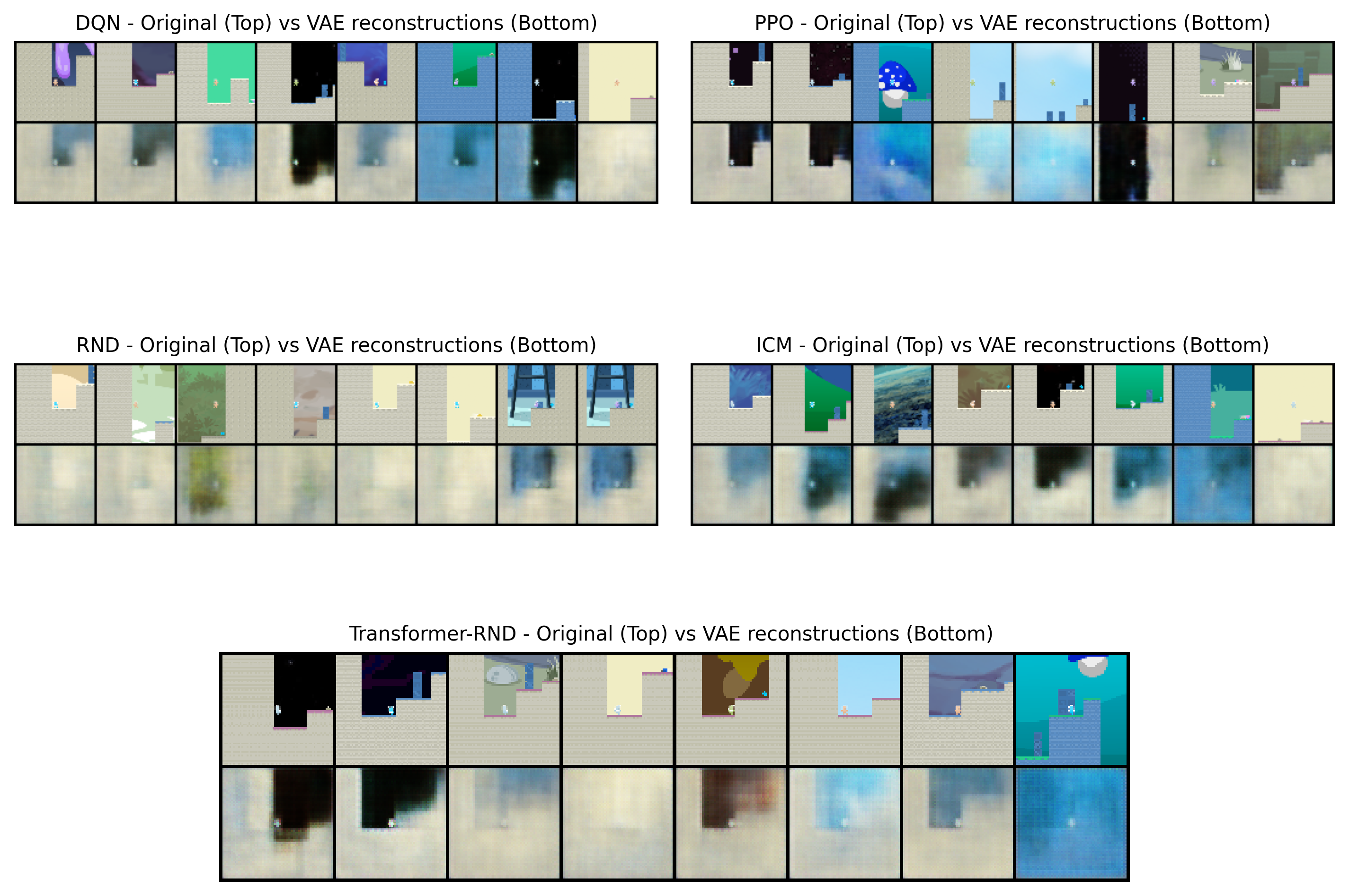}
    \caption{VAE reconstructions for all agents. Each block shows original observations (top) and reconstructions (bottom). ICM and Transformer-RND preserve structural detail, while DQN fails to capture coherent semantics.}
    \label{fig:vae_recon_agents}
\end{figure}

To assess how well each agent encodes environment states, we train a Variational Autoencoder (VAE) separately on the observation datasets collected from each agent. Figure~\ref{fig:vae_recon_agents} presents VAE reconstructions for all five agents. The top rows show original frames, while the bottom rows show corresponding reconstructions.

Qualitative inspection reveals that agents like ICM and Transformer-RND produce high-fidelity reconstructions that preserve scene layout, object boundaries, and agent positions—indicating strong internal state representations. In contrast, DQN reconstructions are blurry and semantically degraded, suggesting underdeveloped or noisy latent encodings. PPO and RND fall in between, showing moderate reconstruction clarity.

\subsection{Latent Space Visualization}

\begin{figure}[h]
    \centering
    \includegraphics[width=0.9\linewidth]{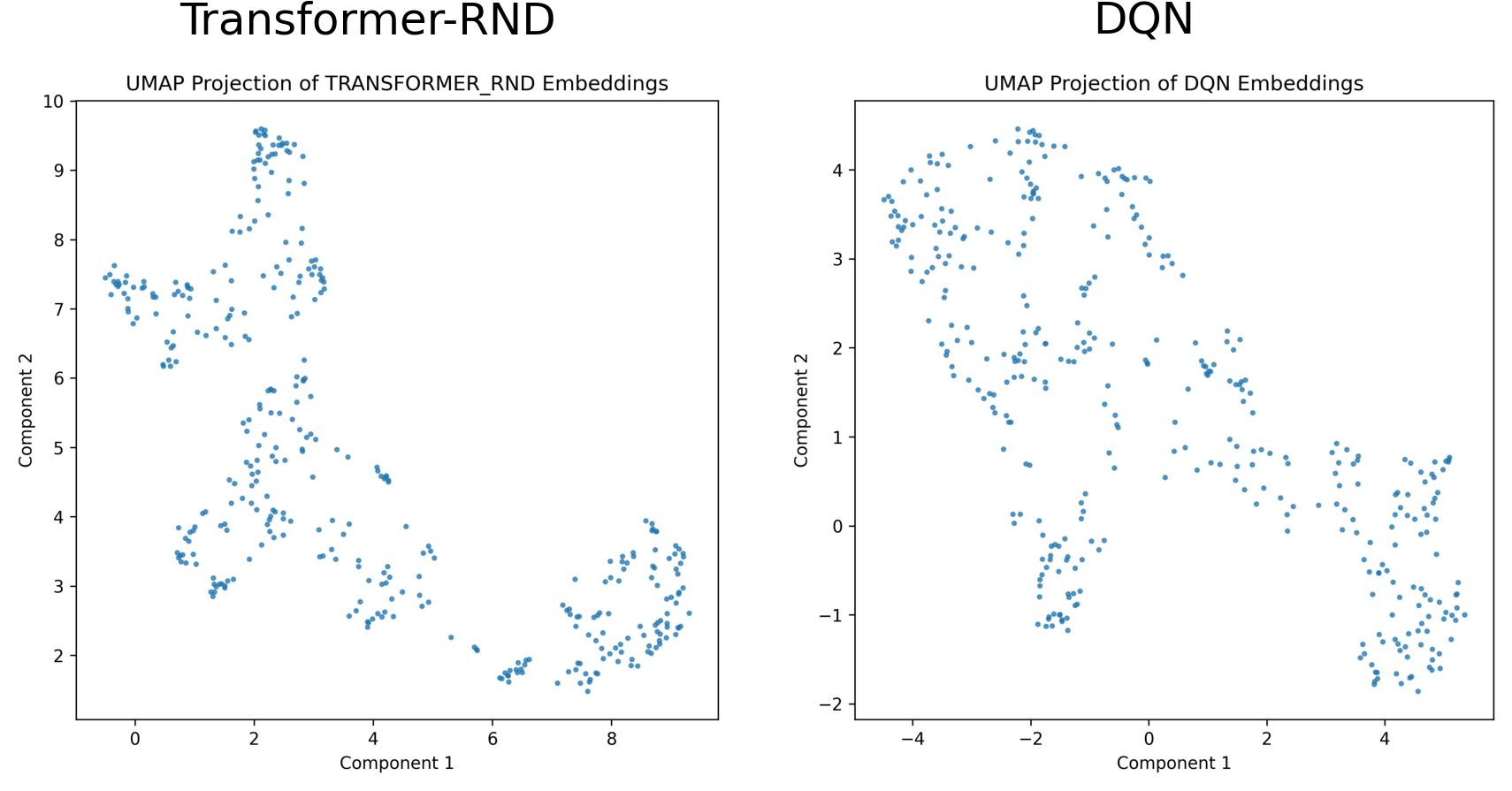}
    \caption{UMAP projection comparison. Transformer-RND (left) forms structured, clustered representations; DQN (right) shows disorganized, overlapping embeddings.}
    \label{fig:umap_best_worst}
\end{figure}

We use UMAP to visualize agent embeddings and assess how effectively each model organizes its internal representations. Figure~\ref{fig:umap_best_worst} compares the Transformer-RND agent, which exhibits strong exploratory behavior, with DQN, which performs poorly. Transformer-RND displays compact, well-separated clusters—indicating meaningful abstraction in latent space. In contrast, DQN’s projection is diffuse and entangled, suggesting weak semantic structuring.

\subsection{Quantitative Metrics}

We report interpretability metrics (Table~\ref{tab:metrics}) including Grad-CAM coverage, attention spread, trajectory entropy, and clustering scores. ICM outperforms in all categories, followed by Transformer-RND. DQN exhibits lowest entropy and clustering quality.

\begin{table}[h]
\centering
\begin{tabular}{lcccc}
\toprule
\textbf{Agent} & \textbf{Coverage (\%)} & \textbf{Entropy} & \textbf{Silhouette} & \textbf{Davies-Bouldin} \\
\midrule
ICM              & 37.6 & 4.91 & 0.46 & 0.83 \\
Transformer-RND  & 35.2 & 4.67 & 0.42 & 0.91 \\
RND              & 33.0 & 4.38 & 0.39 & 1.02 \\
PPO              & 31.5 & 4.01 & 0.34 & 1.18 \\
DQN              & 28.1 & 3.73 & 0.28 & 1.46 \\
\bottomrule
\end{tabular}
\vspace{0.2 cm}
\caption{Summary of interpretability and clustering metrics. Higher entropy and silhouette, lower Davies-Bouldin scores indicate better structure.}
\label{tab:metrics}
\end{table}

\subsection{Experimental Setup}

All models were trained on a single NVIDIA RTX 4060 GPU (8GB VRAM) with 16GB RAM. Each agent was trained for 1M steps, taking approximately 8–9 hours. All analysis was conducted on the same system to ensure consistency.

\vspace{0.5em}
For full visualizations of Grad-CAM, LRP, and latent projections across all agents, refer to the Appendix.

\section{Discussion}

Our experiments provide new insights into how curiosity-driven learning, architectural design, and policy optimization strategies shape the internal mechanisms of unsupervised reinforcement learning agents. By combining visual explanations with quantitative metrics, we interpret how agents perceive, explore, and internally represent their environments.

\subsection{Curiosity Enhances Perceptual Breadth and Representation Quality}

Agents equipped with intrinsic motivation (ICM, RND, Transformer-RND) consistently exhibit broader spatial attention, higher exploration coverage, and richer latent representations than extrinsically motivated agents like DQN. Attention maps and trajectory entropy confirm that curiosity encourages agents to attend to a wider set of features and engage in more diverse behaviors. This broader sensory and behavioral footprint strongly correlates with improved representation quality, as evidenced by higher clustering scores and more structured latent spaces.

\subsection{Transformer Architectures Improve Attention and Abstraction}

The Transformer-RND agent stands out for its ability to balance wide attention coverage with compact, structured latent representations. This suggests that self-attention mechanisms not only enable more expressive perceptual modeling but also support more effective abstraction. In addition to outperforming other agents on exploration coverage, Transformer-RND achieves the highest Calinski-Harabasz clustering score and the lowest attention spread. These results position Transformer-based policies as a promising direction for building interpretable and generalizable RL agents.

\subsection{On-Policy vs. Off-Policy Learning Dynamics}

On-policy agents (PPO, Transformer-RND) exhibit more dynamic attention shifts over time compared to their off-policy counterparts. Higher attention change rates suggest greater perceptual adaptability, likely driven by the tight feedback loop between policy updates and data collection. However, PPO alone demonstrates weaker representation quality, indicating that on-policy learning may require architectural enhancements or stronger intrinsic rewards to support robust abstraction.

\subsection{When Quantitative Metrics Disagree with Intuition}

Interestingly, DQN—despite being the least exploratory—achieves competitive clustering metrics in some cases (e.g., a high Calinski-Harabasz score). We hypothesize that this results from narrow, overfitted representations that are compact but lack semantic richness. This underscores a key takeaway: not all clustering metrics capture generalizability, and interpretability assessments should be triangulated using visualizations and behavioral analysis to yield robust conclusions.

\subsection{Implications and Broader Impact}

The framework and metrics we propose can be used to diagnose policy collapse, monitor exploration health, or identify representational bottlenecks in real-time. Our findings reinforce the importance of interpretability in unsupervised agents, whose behaviors are guided by internally generated signals. As RL progresses toward open-ended learning and real-world deployment, tools for explaining agent behavior will be essential for safety, debugging, and human alignment.

\section{Conclusion}

In this work, we presented a unified interpretability framework to analyze how different reinforcement learning agents perceive, explore, and represent their environments. By combining visual attribution methods (Grad-CAM, LRP) with latent representation analysis via a Variational Autoencoder, we systematically compared five agents—DQN, PPO, ICM, RND, and Transformer-RND—trained on the CoinRun environment.

Our findings show that curiosity-driven agents, particularly ICM and Transformer-RND, exhibit stronger attention development, richer state exploration, and more structured latent embeddings. In contrast, value-based agents like DQN demonstrate diffuse attention and unstructured internal representations, correlating with lower performance and limited exploration. Visual tools such as Grad-CAM and LRP revealed agents' evolving focus, while clustering metrics and VAE reconstructions quantified the organization of their learned state spaces. Notably, our custom metrics—\textit{attention diversity} and \textit{attention change rate}—proved effective in capturing spatial and temporal patterns of agent perception.

This study highlights the role of architectural biases and training signals in shaping internal dynamics, and underscores the importance of interpretability in unsupervised RL. Understanding how agents form internal models is essential for building transparent, robust AI systems, especially in real-world domains such as robotics or autonomous driving.

\paragraph{Limitations.} Our analysis focuses on a single environment and fixed hyperparameters. While the observed trends are consistent and reproducible, future work should test generalizability across tasks and agent configurations.

\paragraph{Future Work.} This analysis opens several promising directions:
\begin{itemize}
    \item Extending interpretability methods to model-based and hierarchical RL agents for richer reasoning insights.
    \item Incorporating temporal attention visualization in Transformer-based agents to interpret long-term dependencies.
    \item Developing online tools for real-time policy debugging using saliency and latent diagnostics during training.
\end{itemize}

\section*{References}
\medskip
{
\small

[1] Selvaraju, R.R., Cogswell, M., Das, A., Vedantam, R., Parikh, D.\ \& Batra, D.\ (2017) Grad-CAM: Visual Explanations from Deep Networks via Gradient-based Localization. In {\it Proceedings of the IEEE International Conference on Computer Vision (ICCV)}, pp.\ 618–626.

[2] Bach, S., Binder, A., Montavon, G., Klauschen, F., Müller, K.R.\ \& Samek, W.\ (2015) On Pixel-Wise Explanations for Non-Linear Classifier Decisions by Layer-Wise Relevance Propagation. {\it PLoS ONE} {\bf 10}(7):e0130140.

[3] Pathak, D., Agrawal, P., Efros, A.A.\ \& Darrell, T.\ (2017) Curiosity-Driven Exploration by Self-Supervised Prediction. In {\it Proceedings of the IEEE Conference on Computer Vision and Pattern Recognition Workshops (CVPRW)}, pp.\ 16–17.

[4] Burda, Y., Edwards, H., Storkey, A.\ \& Klimov, O.\ (2019) Exploration by Random Network Distillation. In {\it International Conference on Learning Representations (ICLR)}.

[5] Mnih, V., Kavukcuoglu, K., Silver, D., Rusu, A.A., Veness, J., Bellemare, M.G., et al.\ (2015) Human-level control through deep reinforcement learning. {\it Nature} {\bf 518}, 529–533.

[6] Schulman, J., Wolski, F., Dhariwal, P., Radford, A.\ \& Klimov, O.\ (2017) Proximal Policy Optimization Algorithms. {\it arXiv preprint arXiv:1707.06347}.

[7] Vaswani, A., Shazeer, N., Parmar, N., Uszkoreit, J., Jones, L., Gomez, A.N., et al.\ (2017) Attention is All You Need. In {\it Advances in Neural Information Processing Systems 30}, pp.\ 5998–6008.

[8] Cobbe, K., Hesse, C., Hilton, J.\ \& Schulman, J.\ (2020) Leveraging Procedural Generation to Benchmark Reinforcement Learning. In {\it Proceedings of the 37th International Conference on Machine Learning (ICML)}, pp.\ 2048–2056.

[9] van der Maaten, L.\ \& Hinton, G.\ (2008) Visualizing Data using t-SNE. {\it Journal of Machine Learning Research} {\bf 9}, 2579–2605.

[10] McInnes, L., Healy, J.\ \& Melville, J.\ (2018) UMAP: Uniform Manifold Approximation and Projection for Dimension Reduction. {\it arXiv preprint arXiv:1802.03426}.

[11] Rousseeuw, P.J.\ (1987) Silhouettes: A graphical aid to the interpretation and validation of cluster analysis. {\it Journal of Computational and Applied Mathematics} {\bf 20}, 53–65.

[12] Davies, D.L.\ \& Bouldin, D.W.\ (1979) A Cluster Separation Measure. {\it IEEE Transactions on Pattern Analysis and Machine Intelligence} {\bf PAMI-1}(2), 224–227.

[13] Calinski, T.\ \& Harabasz, J.\ (1974) A Dendrite Method for Cluster Analysis. {\it Communications in Statistics} {\bf 3}(1), 1–27.


\appendix

\section{Appendix / supplemental material}
This appendix contains additional visualizations and analysis that support the results discussed in the main paper.

\subsection*{A. Grad-CAM and LRP Visualizations}

For each agent, we show Grad-CAM and LRP heatmaps at key training checkpoints (0k, 100k, 500k, 800k, 995k). These visualizations highlight how attention evolves over time and help assess where each agent focuses during policy learning.

\begin{figure}[h]
\centering
\includegraphics[width=0.95\linewidth]{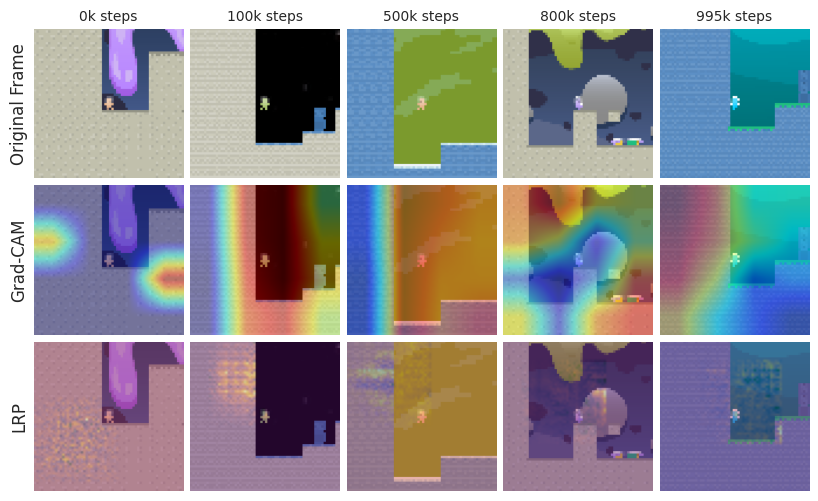}
\caption{DQN: Grad-CAM and LRP visualizations across training steps.}
\end{figure}

\begin{figure}[h]
\centering
\includegraphics[width=0.95\linewidth]{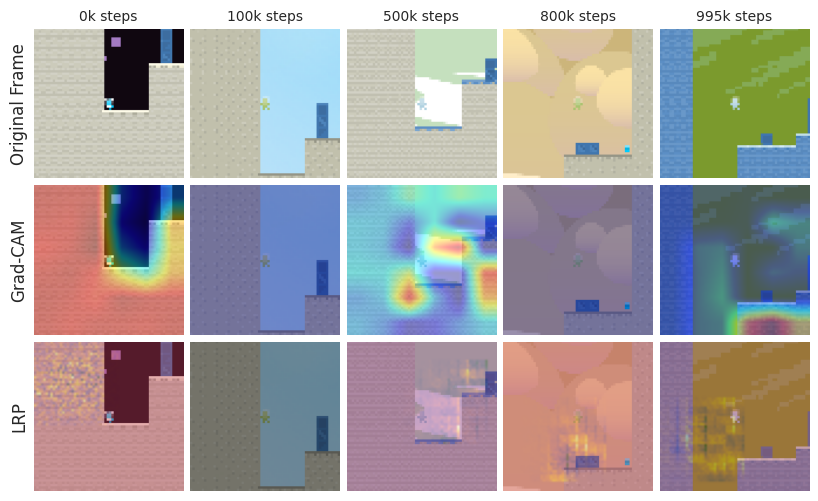}
\caption{PPO: Grad-CAM and LRP visualizations across training steps.}
\end{figure}

\begin{figure}[h]
\centering
\includegraphics[width=0.95\linewidth]{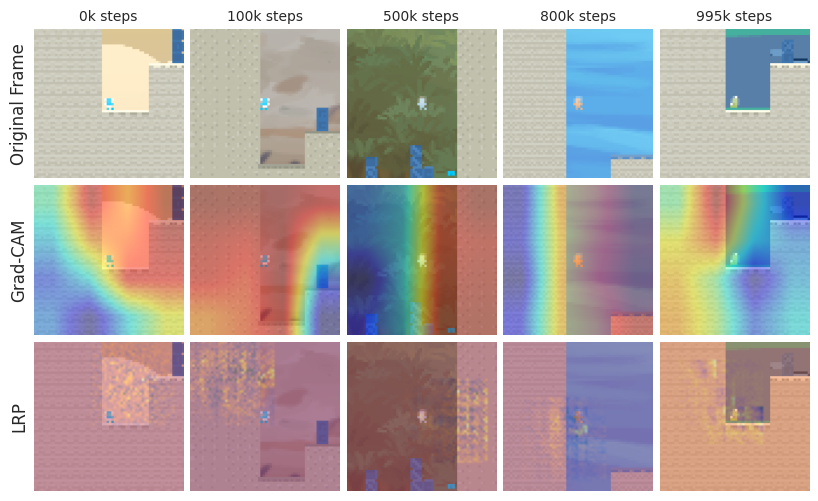}
\caption{RND: Grad-CAM and LRP visualizations across training steps.}
\end{figure}

\begin{figure}[h]
\centering
\includegraphics[width=0.95\linewidth]{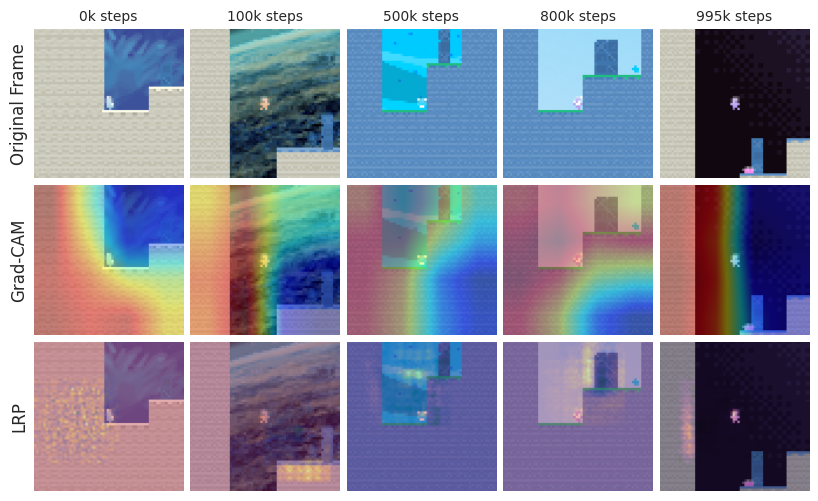}
\caption{ICM: Grad-CAM and LRP visualizations across training steps.}
\end{figure}

\begin{figure}[h]
\centering
\includegraphics[width=0.95\linewidth]{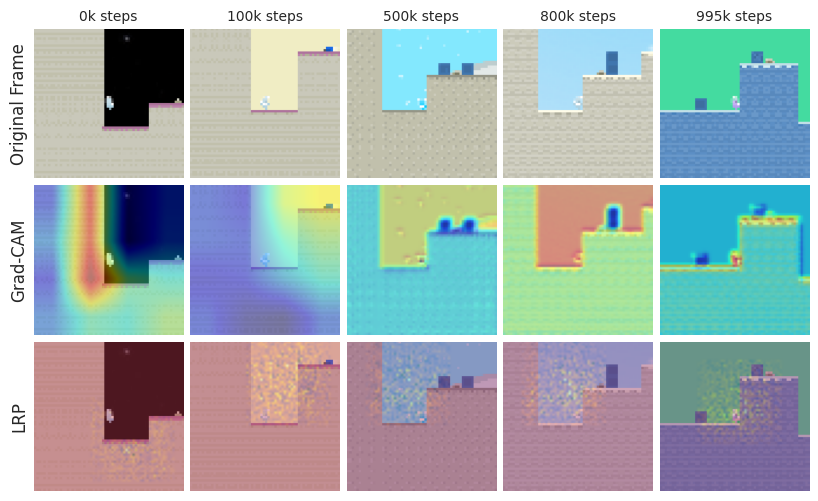}
\caption{Transformer-RND: Grad-CAM and LRP visualizations across training steps.}
\end{figure}

\subsection*{B. Latent Space Projections}

We include both UMAP and t-SNE projections of VAE latent embeddings for each agent. These visualizations reflect the structure and diversity of internal representations formed during training. UMAP emphasizes global structure, while t-SNE highlights local neighborhood preservation.

\begin{figure}[h]
\centering

\begin{subfigure}{0.48\linewidth}
    \includegraphics[width=\linewidth]{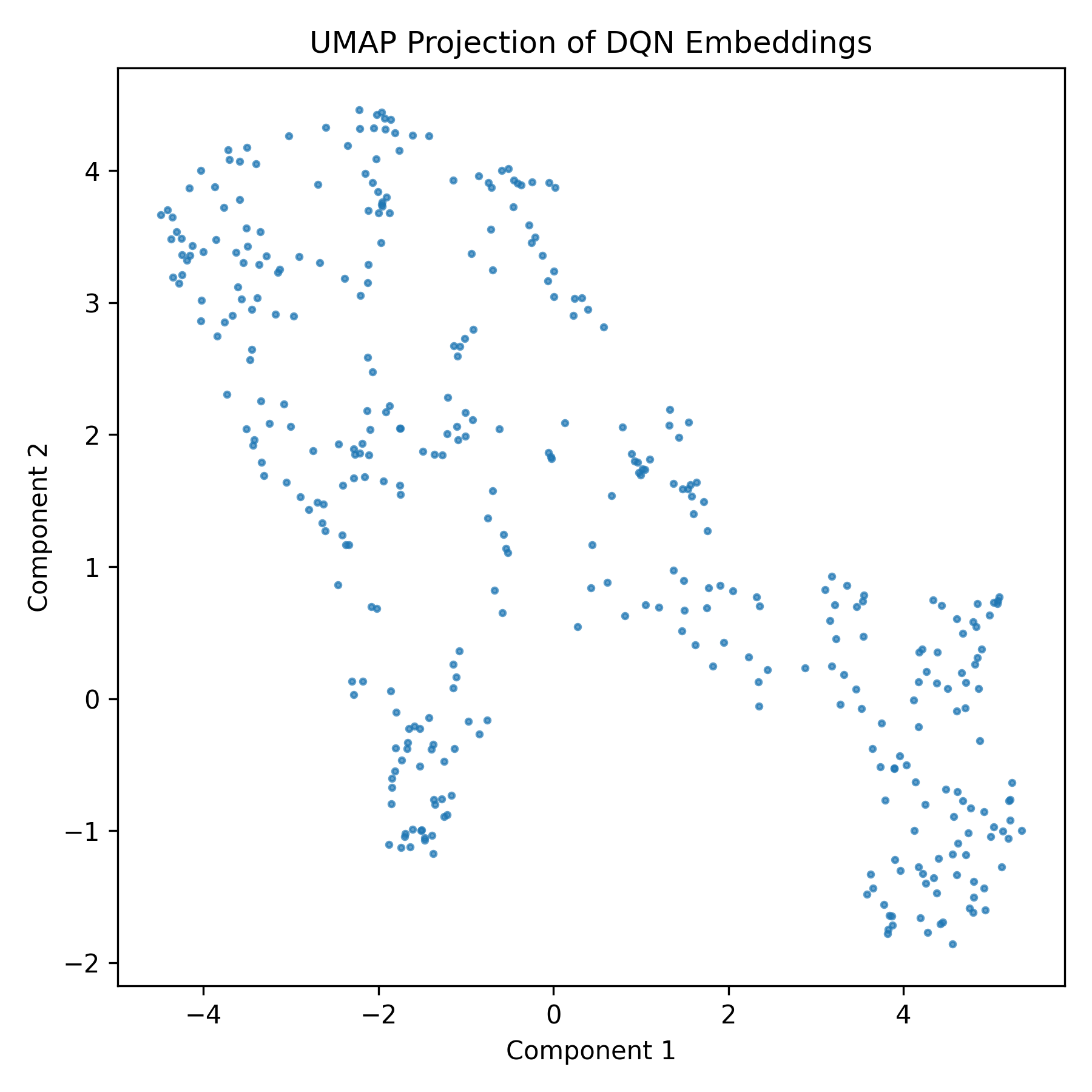}
    \caption*{DQN – UMAP}
\end{subfigure}
\hfill
\begin{subfigure}{0.48\linewidth}
    \includegraphics[width=\linewidth]{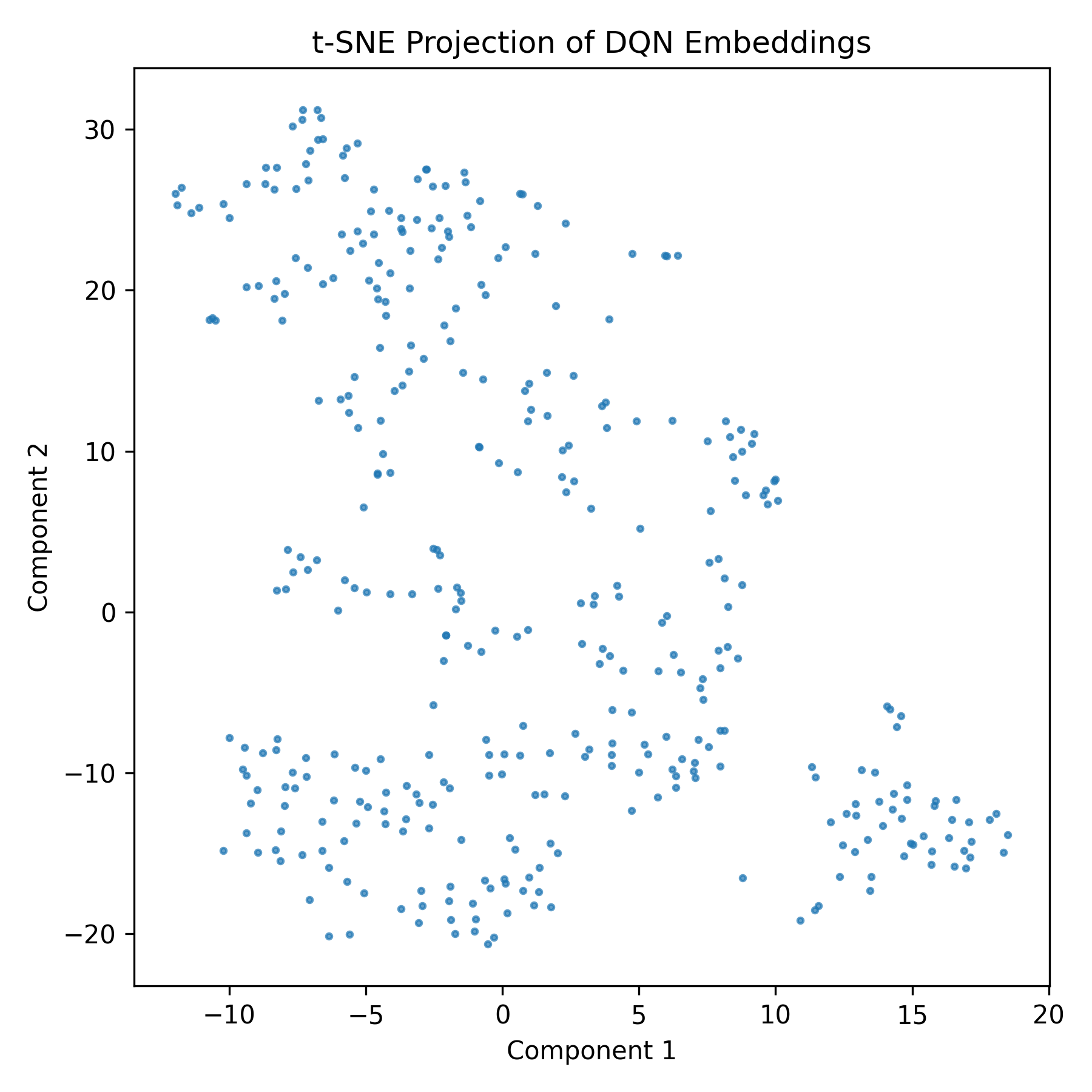}
    \caption*{DQN – t-SNE}
\end{subfigure}

\vspace{1em}

\begin{subfigure}{0.48\linewidth}
    \includegraphics[width=\linewidth]{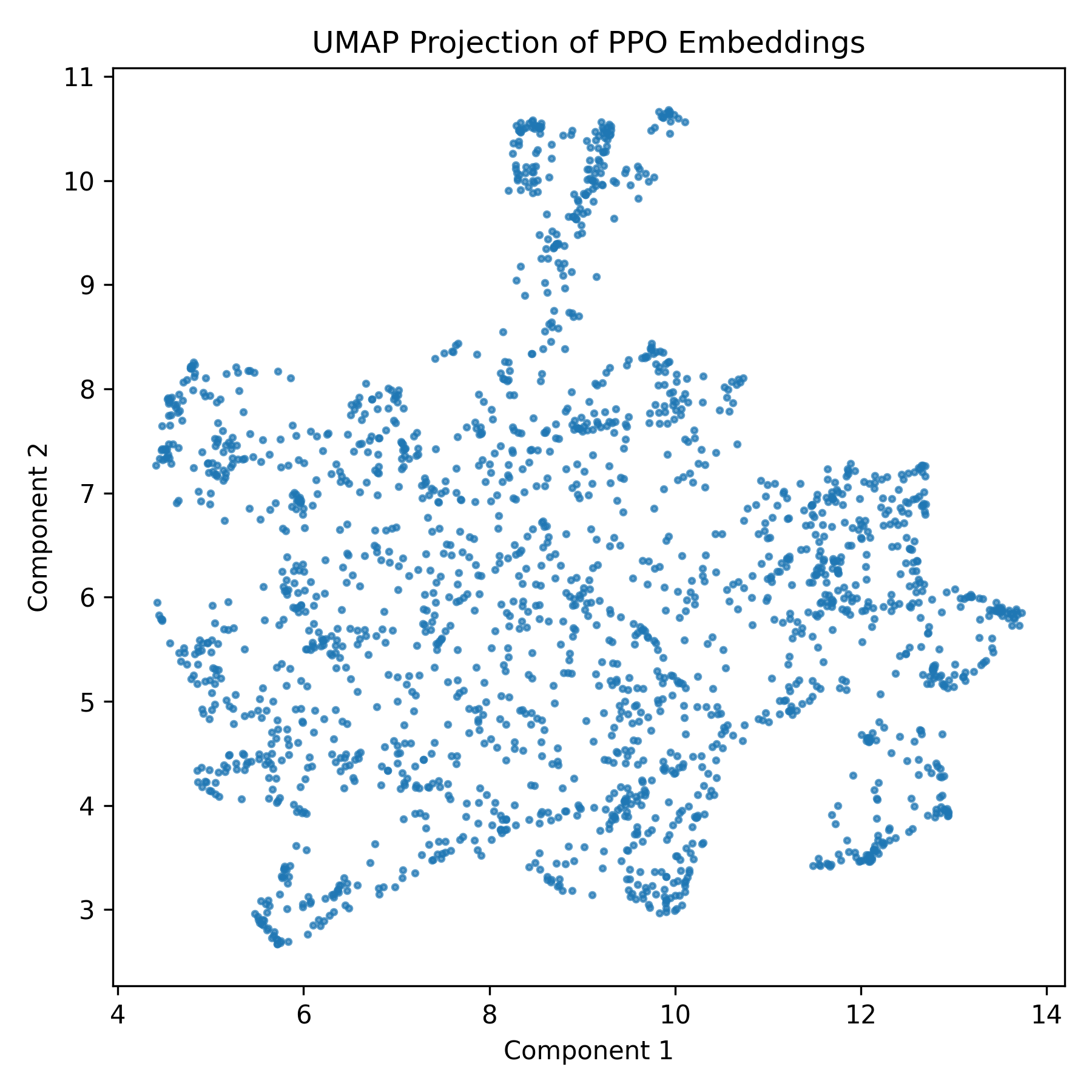}
    \caption*{PPO – UMAP}
\end{subfigure}
\hfill
\begin{subfigure}{0.48\linewidth}
    \includegraphics[width=\linewidth]{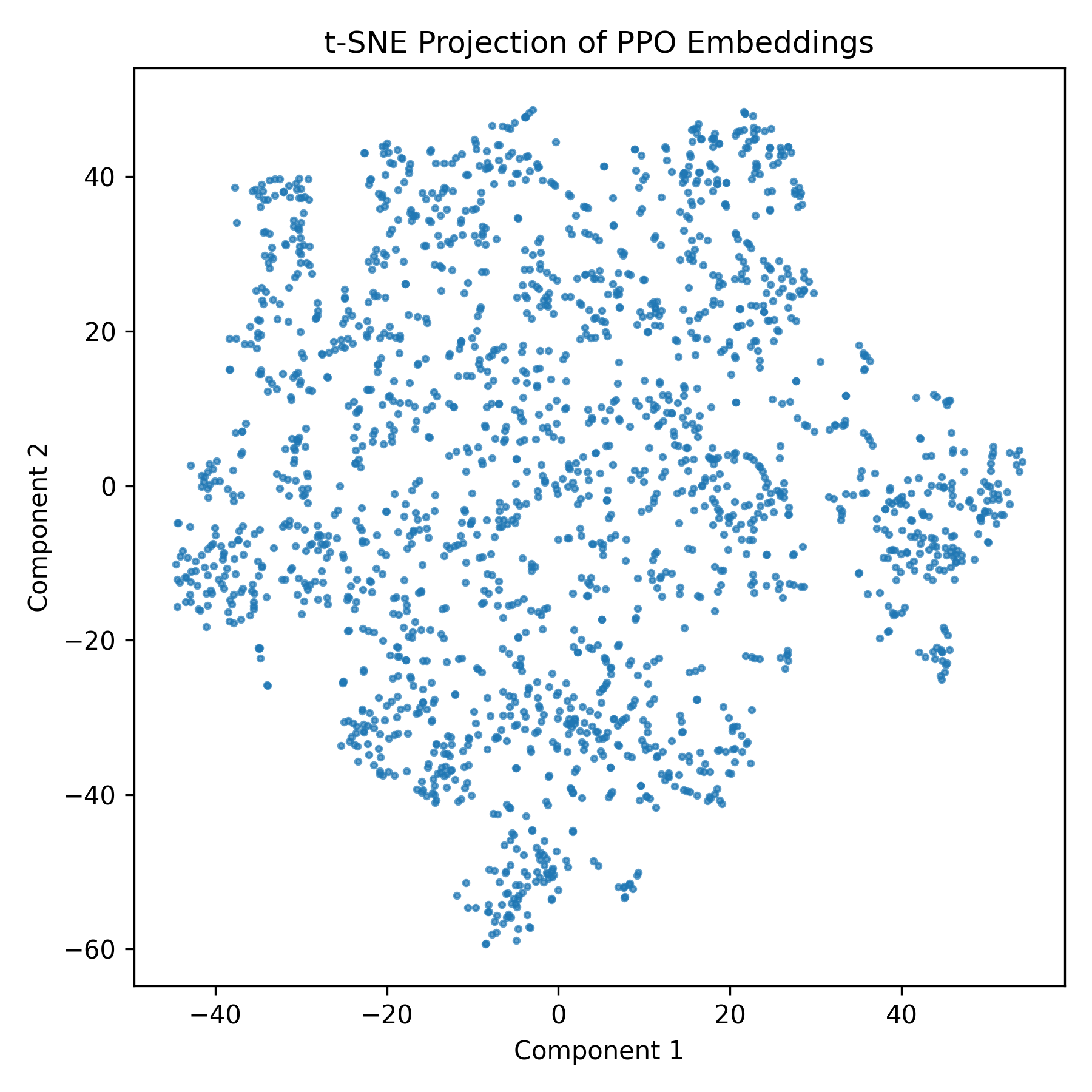}
    \caption*{PPO – t-SNE}
\end{subfigure}

\vspace{1em}

\begin{subfigure}{0.48\linewidth}
    \includegraphics[width=\linewidth]{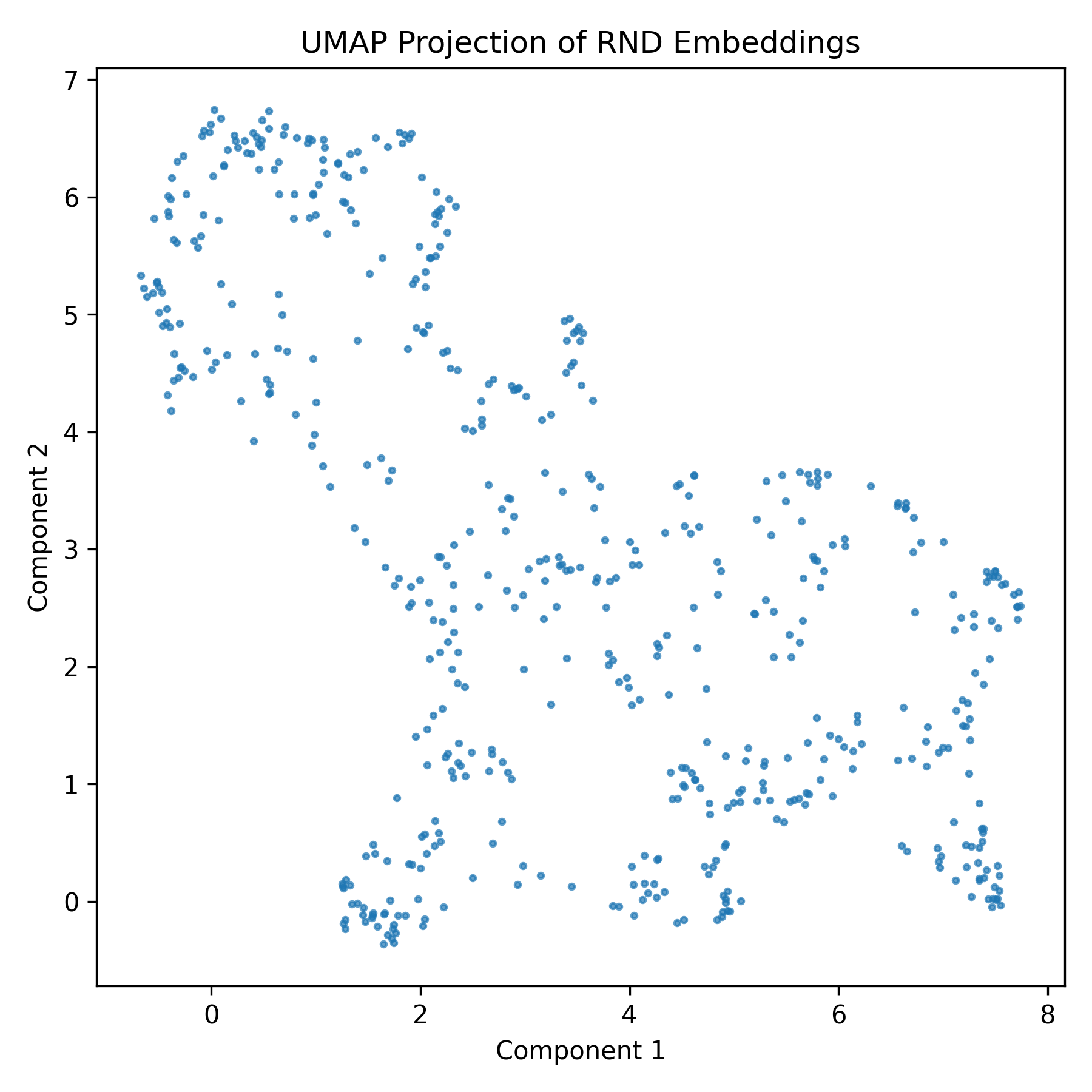}
    \caption*{RND – UMAP}
\end{subfigure}
\hfill
\begin{subfigure}{0.48\linewidth}
    \includegraphics[width=\linewidth]{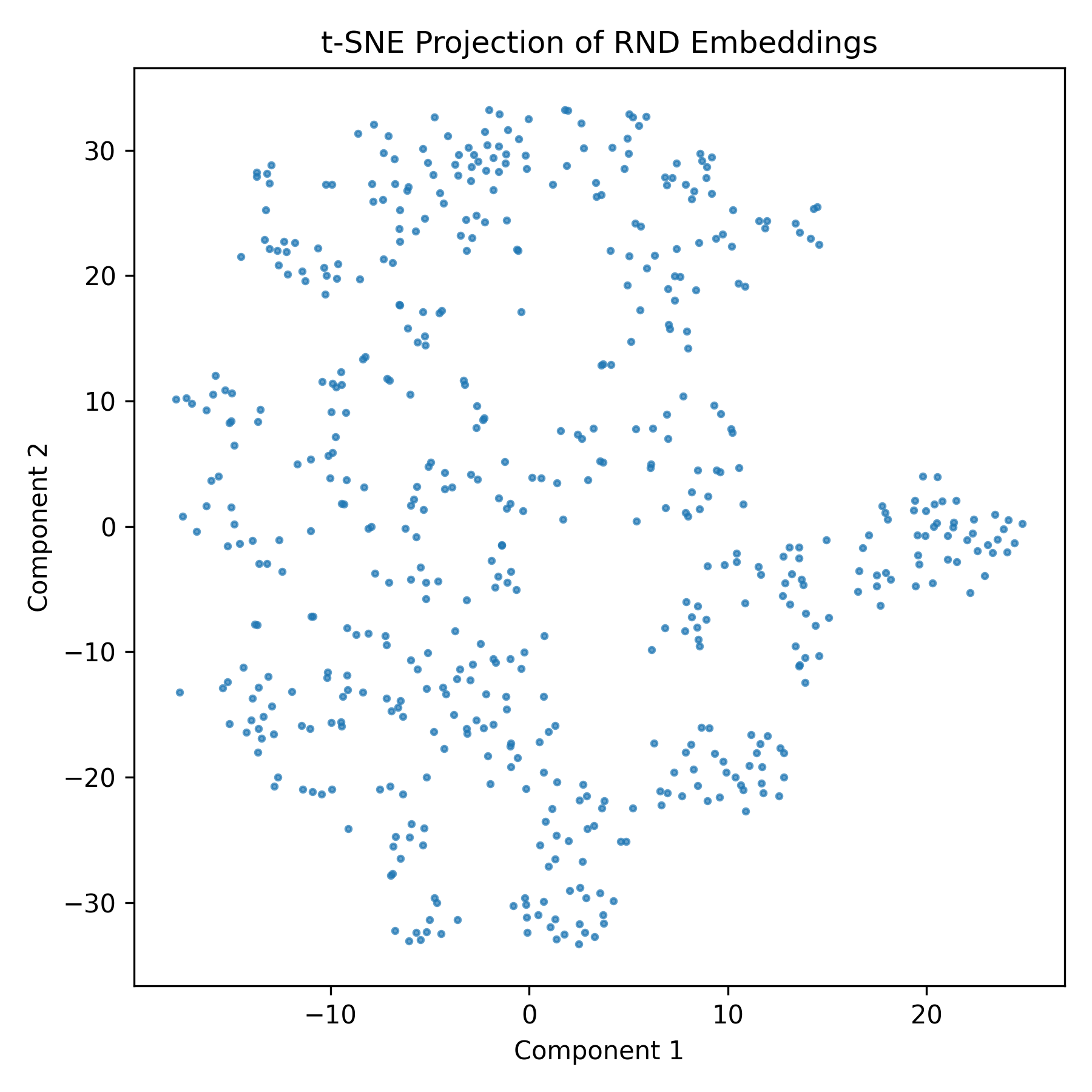}
    \caption*{RND – t-SNE}
\end{subfigure}

\caption{UMAP and t-SNE projections of latent embeddings for DQN, PPO, and RND.}
\label{fig:latent_projections_1}
\end{figure}

\begin{figure}[h]
\centering

\begin{subfigure}{0.48\linewidth}
    \includegraphics[width=\linewidth]{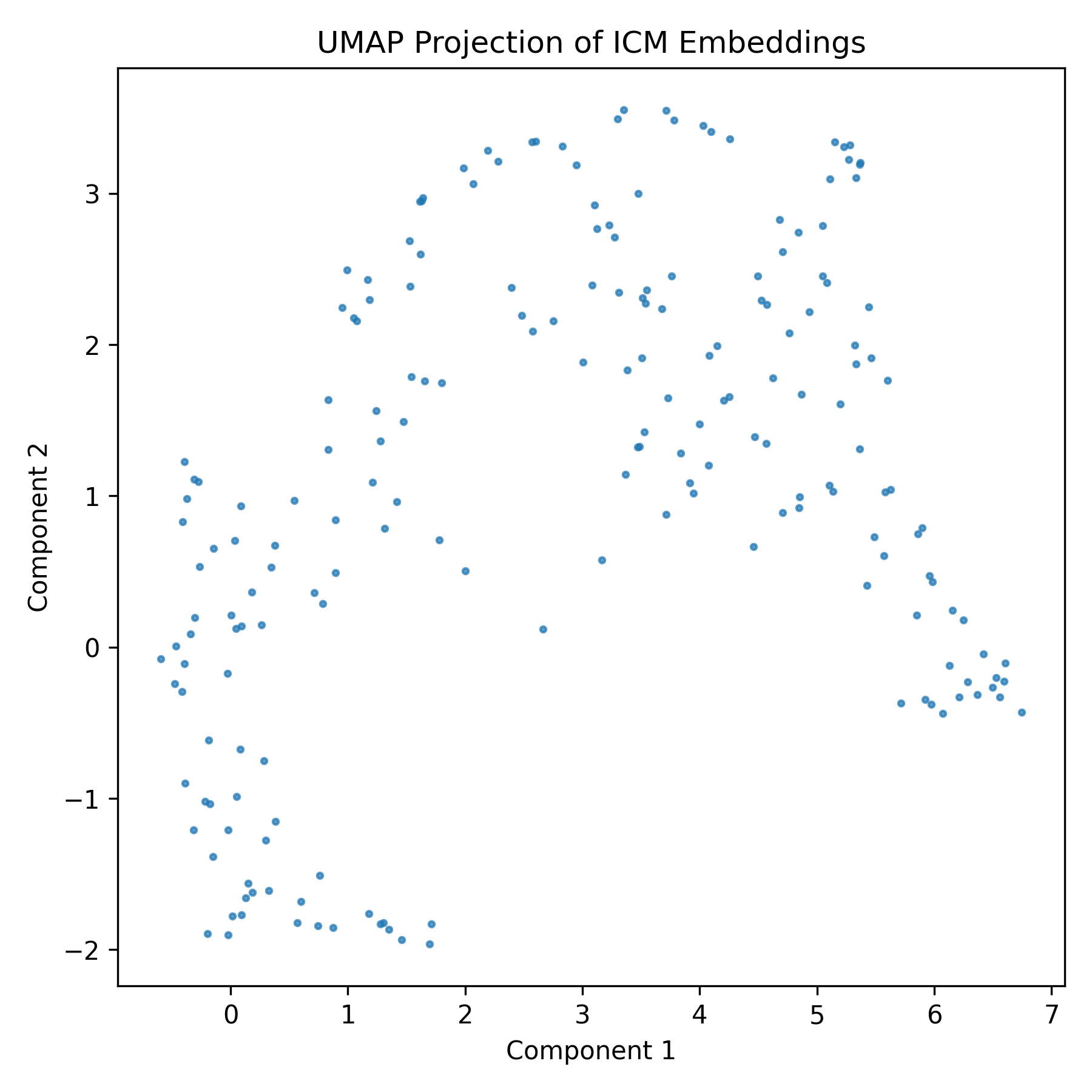}
    \caption*{ICM – UMAP}
\end{subfigure}
\hfill
\begin{subfigure}{0.48\linewidth}
    \includegraphics[width=\linewidth]{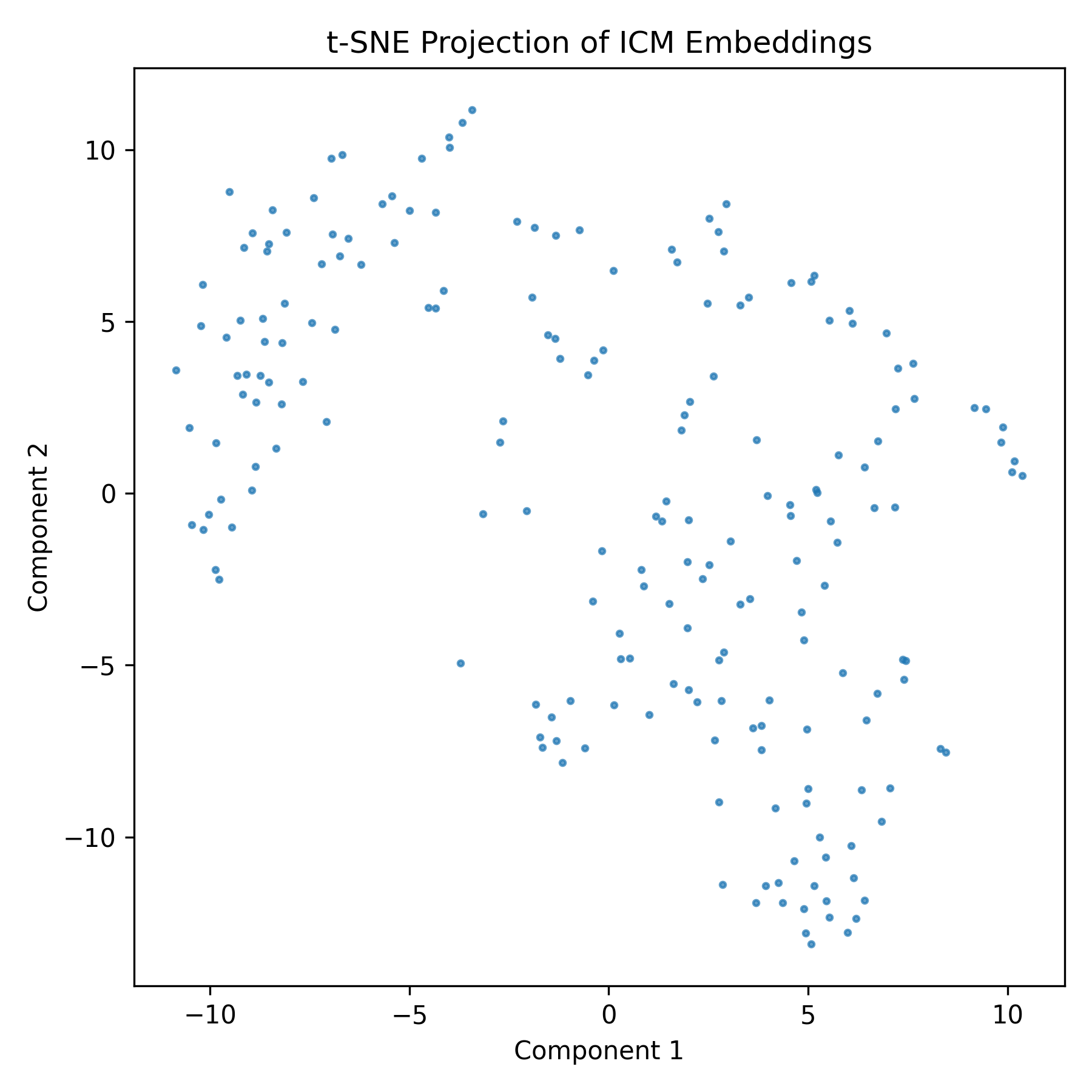}
    \caption*{ICM – t-SNE}
\end{subfigure}

\vspace{1em}

\begin{subfigure}{0.48\linewidth}
    \includegraphics[width=\linewidth]{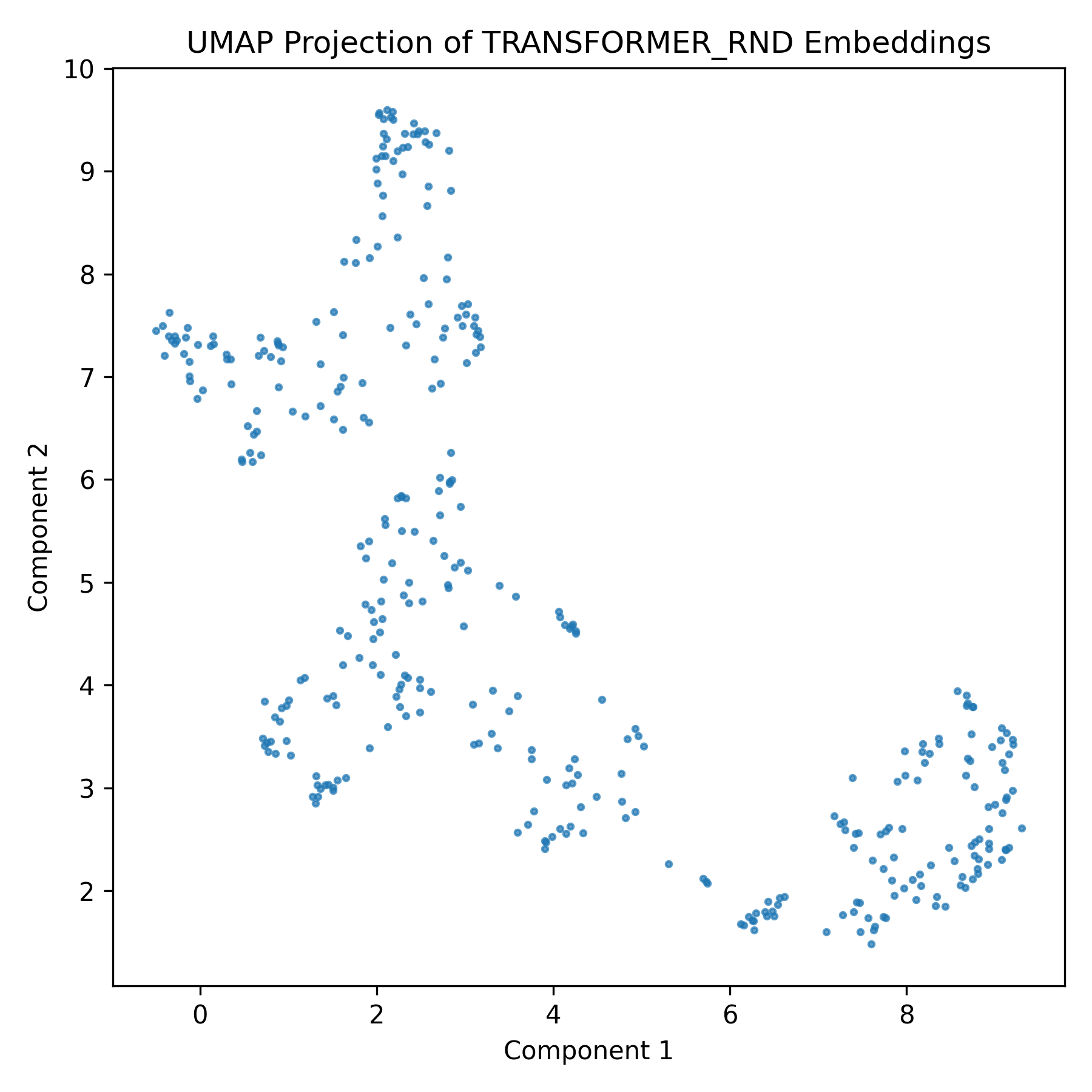}
    \caption*{Transformer-RND – UMAP}
\end{subfigure}
\hfill
\begin{subfigure}{0.48\linewidth}
    \includegraphics[width=\linewidth]{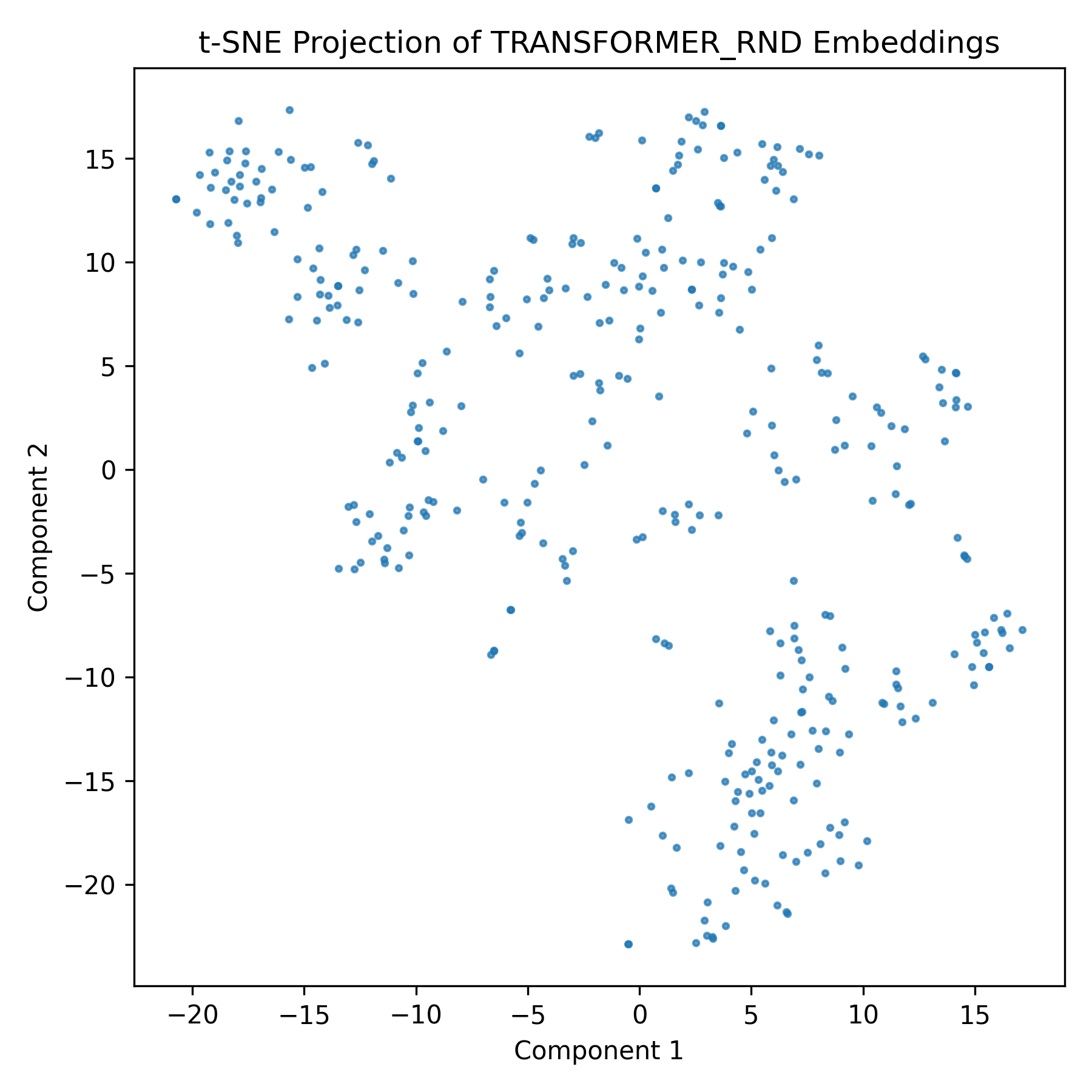}
    \caption*{Transformer-RND – t-SNE}
\end{subfigure}

\caption{UMAP and t-SNE projections of latent embeddings for ICM and Transformer-RND.}
\label{fig:latent_projections_2}
\end{figure}

\subsection*{C. Compute Infrastructure}
All experiments were conducted on a single NVIDIA RTX 4060 (8GB VRAM) with 16GB system RAM. Each agent required 8–9 hours to train for 1M steps. All VAE and interpretability analyses were run post-training on the same hardware.

\end{document}